\documentclass[10pt,twocolumn,letterpaper]{article}

\usepackage{cvpr}
\usepackage{times}
\usepackage{epsfig}
\usepackage{graphicx}
\usepackage{amsmath}
\usepackage{amssymb}
\usepackage[normalem]{ulem}
\usepackage{booktabs} 
\usepackage{paralist} 
\usepackage{subfig}
\usepackage{tabularx}
\usepackage[table]{xcolor} 
\usepackage{paralist}

\usepackage{threeparttable}
\usepackage{algorithm}
\usepackage[noend]{algpseudocode}

\usepackage{mathrsfs}

\usepackage[pagebackref=true,breaklinks=true,letterpaper=true,colorlinks,bookmarks=false]{hyperref}

\cvprfinalcopy 

\def\cvprPaperID{4157} 

\ifcvprfinal\fi
\begin{document}


\title{On Finding Gray Pixels}

\author{Yanlin Qian$^{1,3}$, Joni-Kristian K\"am\"ar\"ainen$^1$, Jarno Nikkanen$^2$, Jiri Matas$^{1,3}$\\
  $^1$Computing Sciences, Tampere University~~
  $^2$Intel Finland\\
  $^3$Center for Machine Perception, Czech Technical University in Prague\\
}

\maketitle
\thispagestyle{empty}


\begin{abstract}
We propose a novel grayness index for finding gray pixels and demonstrate its effectiveness and efficiency in illumination estimation. 
The grayness index, GI in short, is derived using the Dichromatic Reflection Model and is learning-free. GI allows to estimate one or multiple illumination sources in color-biased images.
On standard single-illumination and multiple-illumination estimation benchmarks, GI outperforms state-of-the-art statistical methods and many recent deep methods. GI is simple and fast, written in a few dozen lines of code, processing a 1080p image in $\sim 0.4$ seconds with a non-optimized Matlab code.
 
\end{abstract}


\section{Introduction}
\begin{figure}[t]
\begin{center}
\includegraphics[height=0.40\linewidth,width=0.99\linewidth]{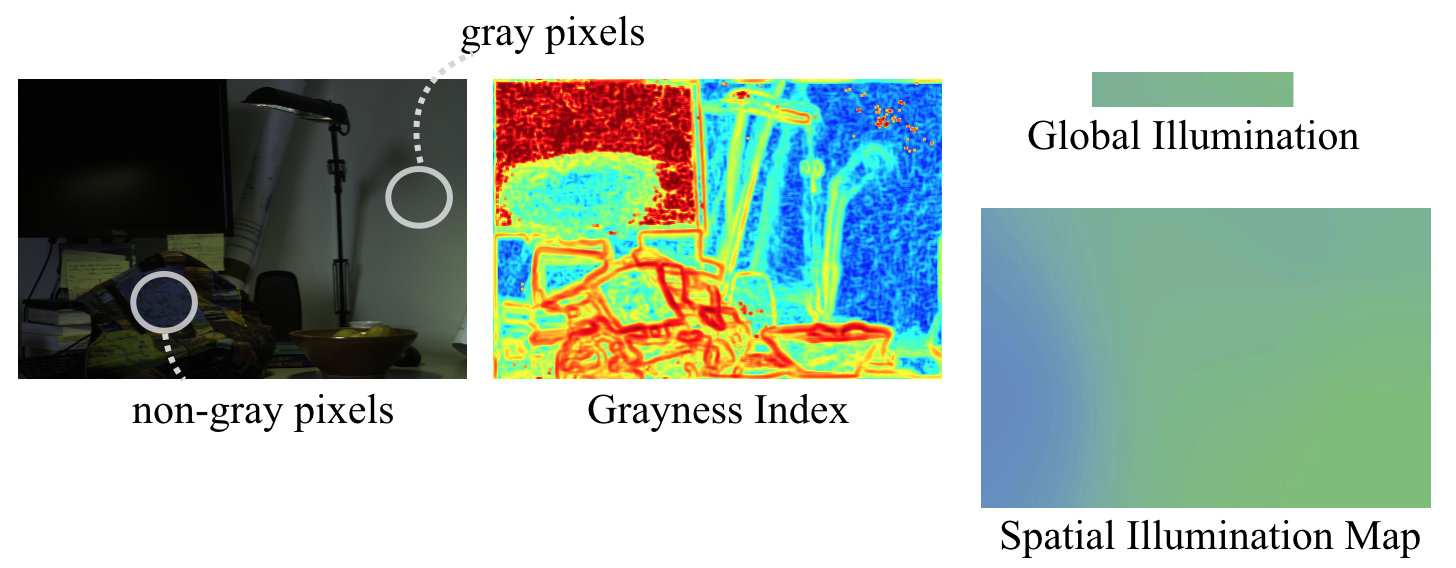}
\caption{ 
Gray and non-gray image pixels (left). The Grayness Index (GI) map (middle, blue denotes high grayness value).  
The global (top right) and spatially-variant illumination color (right) estimated from the GI map.
\label{fig:introgi}}
\end{center}\vspace{-7mm}
\end{figure}

The human eye has the ability to adapt to changes in imaging conditions and illumination of scenes. The well-established computer vision problem of \textit{color constancy},  CC in short,
is trying to endow consumer digital cameras with the same ability. With ``perfect'' color constancy, finding a gray pixel is not a problem at all -- just checking whether the RGB values are equal.
However, given a color-biased image, detecting gray pixels, i.e. pixels observing an achromatic surface, is a hard and ill-posed problem  --
imagine a white piece of paper illuminated with a cyan light source; or is it a cyan paper under white light?
On the other hand, ``perfect'' gray pixels in an image indicate that color constancy is satisfied. Thus, from this point onward, we treat \textit{finding gray pixels} and \textit{color constancy} as equivalent problems (see also Fig. \ref{fig:introgi}).
Color constancy problem arises in many computer vision and image processing applications, such as computational photography, intrinsic image decomposition, semantic segmentation, scene rendering, object tracking, \etc ~\cite{foster2011color}.

For decades, \textit{learning-free methods}, the classical approach to color constancy,  have relied on the assumption that the illumination color is constant over the whole scene and can therefore be estimated by  global processing~\cite{brainard1986analysis,barnard2002comparison,van2007edge,finlayson2004shades,gao2014eccv,yang2015efficient,Cheng14}. This approach has the advantage of being independent to the acquisition device, since the illumination properties are estimated on a per-image basis. 
Recently, state-of-the-art \textit{learning-based methods}, including convolutional neural networks (CNNs),  have consistently outperformed statistical methods when validated on specific datasets \cite{chakrabarti2012color,gijsenij2010generalized,gehler2008bayesian,gijsenij2011color,joze2014exemplar}.  
We argue that learning-based methods depend on the assumption that the statistical distribution of the illumination and/or scene content is similar in training and test images. In other words, learning-based methods assume that imaging and illumination conditions of a given image can be inferred from previous training examples, thus becoming heavily dependent on the training data \cite{gao2017josa}.

In this paper, we focus on the learning-free approach. 
For a practical example,  consider the case when a user retrieves a linear-RGB (gamma corrected) image from the web and wants to correct its colors. In this scenario, in which the used CC method has never seen images from that camera, illumination estimation and color correction must be performed without strong assumptions on the imaging device or the captured scene. We experimentally show that in this setting, learning-free methods show more promising and robust results as compared to learning-based methods.
As a result, there is a great need for learning-free approaches that are insensitive to parameters such as the camera and imaging process of captured images.

In most camera sensors, gray pixels are rendered gray in linear-RGB image under standard neutral illumination, making \textit{grayness} a potential measure to estimate the color of incident illumination. We adopt Shafer's Dichromatic Reflection Model (DRM)~\cite{shafer1985using} to develop a novel grayness index (GI), which allow ranking all image pixels according to their grayness. The appealing points are: \textrm{(i)} GI is simple and fast to compute; \textrm{(ii)} it has a clear physical meaning; \textrm{(iii)} it can handle specular highlights to some extend (from qualitative comparison); \textrm{(iv)} it allows pixel-level illumination estimation; \textrm{(v)} it provides consistent prediction across different cameras. Comprehensive results on single-illumination and multi-illumination color constancy datasets show that GI outperforms the state-of-the-art learning-free methods and achieves state-of-the-art in the cross-dataset setting.

\section{Related Work}
Consider image $I$ captured using a linear digital camera sensor, with black level corrected and no saturation. In the dichromatic reflection model, the pixel value at $(x,y)$ under one global illumination source can be modeled as \cite{shafer1985using}: 
\begin{align}
\label{eq:formation}
I_{i}^{(x,y)} =\gamma_{b}^{(x,y)}  \int F_{i}(\lambda) L(\lambda)  R_{b}^{(x,y)}(\lambda) d\lambda \notag \\+ 
\gamma_{s}^{(x,y)} \int F_{i}(\lambda) L(\lambda)  R_{s}^{(x,y)}(\lambda) d\lambda,
\end{align}
where $I_{i}^{(x,y)}$ is the pixel value at $(x,y)$, $L(\lambda)$ the global light spectral distribution, $F_{i}(\lambda)$ the sensor sensitivity, $i=\{R,G,B\}$ for trichromatic cameras, and $\lambda$ the wavelength. The chromatic terms  $R_{b}(\lambda)$ and $R_{s}(\lambda)$ account for body and surface reflection, respectively, while the achromatic terms $\gamma_{b} $ and $\gamma_{s}$ are the intensities of the above two types of reflection.

In addition, under the the assumption of narrow spectral response $F_i(\lambda)$, Eq. \ref{eq:formation} is further simplified to \cite{barron2015convolutional}:

\begin{equation}
I^{(x,y)} =  W ^{(x,y)}  \circ L + V ^{(x,y)}  \circ L , 
\label{eq:simpleformation}
\end{equation}

where $ \circ $ denotes Hadamard Product and,

\begin{align}
W  ^{(x,y)} &=[\gamma_{b}^{(x,y)}  R_{b,R}^{(x,y)},\gamma_{b}^{(x,y)}  R_{b,G}^{(x,y)},\gamma_{b}^{(x,y)}  R_{b,B}^{(x,y)}]^{T}, \notag \\
V  ^{(x,y)} &=[\gamma_{s}^{(x,y)}  R_{s,R}^{(x,y)},\gamma_{s}^{(x,y)}  R_{s,G}^{(x,y)},\gamma_{s}^{(x,y)}  R_{s,B}^{(x,y)}]^{T}, \notag \\ 
L &=[F_R L_R, F_G L_G,F_B L_B]^{T}, 
\label{eq:wvl}
\end{align}

where the \{R,G,B\} subscripts represent the corresponding parts of the spectrum that intersect with $F_i$. 
Eq.~\ref{eq:simpleformation} shows the formation of a pixel value in image $I$ corresponding to a location in the scene exhibiting body $W$ and surface reflection~$V$, under a camera-captured global light~$L$.

The goal of CC is to estimate $L$ in order to recover $W$, given $I$. Based on the strategy used for solving this problem, we divide color constancy methods into two categories: learning-based, learning-free methods.

\paragraph{Learning-based Methods} \cite{chakrabarti2012color,gijsenij2010generalized,gehler2008bayesian,gijsenij2011color,joze2014exemplar,yanlin2016icpr,yanlin2017iccv} aim at building a model that relates the captured image $I$ and the sought illumination $L$ from extensive training data. Among the best-performing state-of-the-art approaches, the CCC method \cite{barron2015convolutional} discriminatively learns convolutional filters in a 2D log-chroma space. This framework was subsequently accelerated using the Fast Fourier Transform on a chroma torus~\cite{barron2017fourier}. Chakrabarti \textit{et al.}~\cite{chakrabarti2015color} leverage the normalized luminance for illumination prediction by learning a conditional chroma distribution.
DS-Net~\cite{shi2016eccv} and FC$^4$ Net~\cite{hu2017cvpr} are two deep learning methods, where the former chooses an estimate from multiple illumination guesses using a two-branch CNN architecture and the later addresses local estimation ambiguities of patches using a segmentation-like framework. 
Learning-based methods achieve great success in predicting pre-recorded ``ground-truth'' illumination color fairly accurately, but heavily depending on the same cameras and/or scenes being in both training and test images (see Sec. \ref{sec:novelgi} and Sec. \ref{sec:agnostic}). The Corrected-Moment method \cite{finlayson2013corrected} can also be considered as a learning-based method as it needs to train a corrected matrix for each dataset. 

\paragraph{Learning-free Methods} 
estimate the illumination by making prior assumptions about the local or global regularity of the illumination and reflectance. The simplest such method is \textit{Gray World} \cite{buchsbaum1980spatial} that assumes that the global average of  reflectance is achromatic. The generalization of this assumption by restricting it to local patches and higher-order gradients has led to more powerful statistics-based methods, such as White Patch \cite{brainard1986analysis}, General Gray World \cite{barnard2002comparison}, Gray Edge \cite{van2007edge}, Shades-of-Gray \cite{finlayson2004shades} and LSRS \cite{gao2014eccv}, among others \cite{Cheng14}. 

Physics-based Methods \cite{tominaga1996multichannel,finlayson2001convex,finlayson2001solving}, estimate illumination from the understanding of the physical process of image formation (\textit{e.g.} the Dichromatic Model), thus being able to model highlights and inter-reflections. Most physics-based methods estimate illumination based on intersection of multiple dichromatic lines, making them work well on toy images and images with only a few surfaces but often failing on natural images \cite{finlayson2001solving}. The latest physics-based method relies on the longest dichromatic line segment assuming that the Phong reflection model holds and an ambient light exists~\cite{sung2018tip}.  Although our method is based on the Dichromatic Model, we classify our approach as \textit{statistical} since the core of the method is finding gray pixels based on some observed image statistics. We refer readers to \cite{gijsenij2011computational} for more details about physics-based methods.

\paragraph{The Closest Methods to GI} are Xiong \textit{et al.} \cite{xiong2007automatic} and Gray Pixel by Yang {\em et al.}~\cite{yang2015efficient}. Xiong \textit{et al.} \cite{xiong2007automatic} method searches for gray surfaces based on a special LIS space, but it is camera-dependent. Gray Pixel~\cite{yang2015efficient} is closest to our work and is therefore outlined in details in Sec.~\ref{sec:novelgi}.

\section{Grayness Index}
\label{sec:novelgi}

We first review the previous Gray Pixel~\cite{yang2015efficient} (derived from the  Lambertian model) in the context of dichromatic reflection model (DRM).

\subsection{Gray Pixel in~\cite{yang2015efficient}}
\label{subsec:oldgp}
Yang {\em et al.}~\cite{yang2015efficient} claims that gray pixels can be sought by a set of constraints. However, their formulation often
identifies gray pixels that clearly are color pixels. This phenomenon has been noticed, but not properly analyzed. Herein we analyze GP using DRM and point out the potential failure cases of the original formulation.

Assuming narrow band sensor, Eq.~\ref{eq:formation} simplifies to:

\vspace{-5mm}
\begin{align}
\small
I_{i}^{(x,y)}\!=\!\gamma_{b}^{(x,y)} F_{i} L_{i}  R_{b,i}^{(x,y)}\!+\!\gamma_{s}^{(x,y)}  F_{i} L_i  R_{s,i}^{(x,y)},\notag \\
\!i\!\in\!\{\mbox{R,G,B}\}. 
\label{eq:simplied_drm}
\end{align}
Then, following Yang~\textit{et al.}~\cite{yang2015efficient}, we apply $log(\cdot)$ and a local constrast operator $C\{\cdot\}$ 
(Laplacian of Gaussian, see Sec.~4 for more details) on the both sides, and obtain

\vspace{-5mm}
\begin{align}
\label{eq:delta_drm}
\small
C\{\log(I_i^{(x,y)})\}  &=C\{\log(F_{i}L_{i}R_{b,i}^{(x,y)})\} \notag \\
+& C \left\lbrace\log\left(\gamma_b^{(x,y)}+\gamma_s^{(x,y)} \frac{R_{s,i}^{(x,y)}}{R_{b,i}^{(x,y)}} \right) \right\rbrace .
\end{align}

\noindent If $\gamma_s=0$ (means no surface reflection), we obtain: 

\vspace{-5mm}
\begin{align}
\small
C\{\log(I_i^{(x,y)})\} = C\{\log(\gamma_b^{(x,y)} R_{b,i}^{(x,y)})\} \enspace . 
\label{eq:delta_lsa}
\end{align}
If $\gamma_s \neq 0$, due to the interaction between $\gamma_b$ and $ \gamma_s \frac{R_{s,i}}{R_{b,i}}$ in Eq. \ref{eq:delta_drm}, those colored pixels can be wrongly identified as gray pixels. 
Central to GP is that, when $\gamma_s=0$, a non-uniform intensity casting on a homogeneous gray surface can induce the same amount of ``contrast'' in each channel. Varying intensity of light may result from the geometry between surface and illumination (shading) and that among different surfaces (occlusion). In order to resolve this problem we adopt the Dichromatic Reflection Model,  exploring another path to identify gray pixels in a more complex environment.

\subsection{Grayness Index using Dichromatic Reflection Model}

For simplicity, in the sequel we will drop the superscripts $(x,y)$, as all operations are applied in a local neighborhood centered at $(x,y)$.
We first calculate the residual of the red channel and luminance in log space and then apply local contrast operator $C\{\cdot\}$ to Eq. \ref{eq:delta_drm} as:

\vspace{-5mm}
\begin{align}
\hspace{-3ex}
\small
 C&\{\log(I_R)\!-\!\log(\lvert I \lvert) \}\!=\!C\{\log(F_{R}L_{R})\!+\!\log(\gamma_b R_{b,R}\!+\!\gamma_s {R_{s,R}} ) \} \notag \\ 
&-\!C\{\log(F_{R}L_{R}(\gamma_b {R_{b,R}}\!+\!\gamma_s {R_{s,R}})\!+\! F_{G}L_{G}(\gamma_b {R_{b,G}}\!+\!\gamma_s {R_{s,G}}) \notag \\
&+\!F_{B}L_{B}(\gamma_b {R_{b,B}}\!+\!\gamma_s {R_{s,B}}) ) \},
\label{eq:delta_residual_drm}
\end{align}
\noindent where $\lvert I \lvert$ denotes the luminance magnitude $(I_R+I_G+I_B)$.

In this case, the neutral interface reflection (NIR) assumption establishes that, for gray pixels, we have that $R_{j,R}=R_{j,G}=R_{j,B}=\bar{R_j}$ with $j\in\{s,b\}$ \cite{lee1990modeling}. In this case, Eq. \ref{eq:delta_residual_drm} simplifies to:

\vspace{-5mm}
\begin{align}
\small
C&\{\log(I_R)\!-\!\log(\lvert I \lvert) \}\!=\!C\{\log(F_{R}L_{R})\!+\!\log(\gamma_b \bar{R_b}\!+\!\gamma_s \bar{R_s} ) \}  \notag \\
&-\!C\{\log ((F_R L_R\!+\!F_G L_G\!+\!F_B L_B)(\gamma_b \bar{R_b}\!+\!\gamma_c \bar{R_s}))\}.   \hspace{-5ex}
\label{eq:delta_residual_graypixels}
\end{align}

In a small local neighborhood, the casting illumination and sensor response can be assumed constant \cite{yang2015efficient}, such that $C\{\log(F_{R}L_{R})\}\!=\!0$ and $C\{\log ((F_R L_R\!+\!F_G L_G\!+\!F_B L_B)\}\!=\!0$, leading to:

\vspace{-5mm}
\begin{equation}
\begin{aligned}
\hspace{-3ex}
C\{\log(I_R)\!-\!\log(\lvert I \lvert) \}\!=\!C\left\lbrace \log \frac{\gamma_b \bar{R_b}\!+\!\gamma_c \bar{R_s}}{\gamma_b \bar{R_b}\!+\!\gamma_c \bar{R_s}} \right\rbrace \!\overset{\mathrm{gray}}{=}\!0.
\end{aligned}
\label{eq:delta_residual_zero}
\end{equation}
 
Eq. \eqref{eq:delta_residual_zero} is a necessary yet not a sufficient condition for gray pixels. A more restrictive requirement for the detection of gray pixels is given by extending Eq.~\ref{eq:delta_residual_zero} to one more color channel (using all channels in redundant, the spectral response of R and B rarely overlap in sensors) as:

\vspace{-5mm}
\begin{equation}
\begin{aligned}
C\{\log(I_R)\!-\!\log(\lvert I \lvert) \}\!=\!C\{\log(I_B)\!-\!\log(\lvert I \lvert) \}\!=0. \\
\end{aligned}
\label{eq:constrain_gray_drm}
\end{equation}

From Eq.~\eqref{eq:delta_residual_drm}, we define the grayness index \wrt $I(x,y)$ as: 

\vspace{-5mm}
\begin{align}
GI(x,y)=\Vert [C\{\log(I_R)-\log(\lvert I \lvert) \},\notag \\
C\{\log(I_B)-\log(\lvert I \lvert) \} ] \Vert ,
\label{eq:grayness_newgreypixel}
\end{align}

where $\Vert \cdot \Vert$ refers to the $\ell 2$ norm. The smaller the GI is, the more likely the corresponding pixel is gray.

In addition, we impose a restriction on the local contrast to ensure that a ``small'' GI value comes from grey pixels in varying intensity of light, not a flatten color patch (no spatial cues), written as:

\vspace{-3mm}
\begin{equation}
C\{I_i\} > \epsilon\mbox{, }\forall i\in\lbrace R,G,B \rbrace,
\label{eq:delta_rgbneq0}
\end{equation}
where $\epsilon$ is a small contrast threshold.

\begin{figure}[t]
\begin{center}
\includegraphics[width=0.9\linewidth]{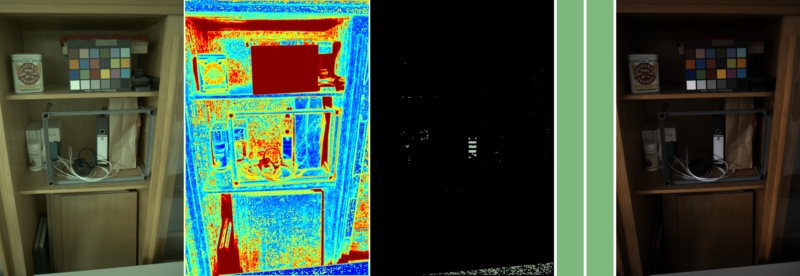}\\
{~~\textit{(a)} ~~~~~~~~~~~~~~ \textit{(b)} ~~~~~~~~~~~  \textit{(c)} ~~~~~ \textit{(d)}\textit{(e)}  ~~~~~  \textit{(f)}}
\caption{ Finding gray pixels.
(a) input image.
(b) computed grayness index $GI$. darker blue indicates higher degree of grayness.
(c) the $N\%$ most gray pixels rendered using the corresponding pixel color (greenish) in (a).
(d) estimated illumination color.
(e) ground truth color.
(f) corrected image using (d).
\label{fig:flowchart}}
\end{center}\vspace{-1cm}
\end{figure} 

The process of computing GI is in two steps:
\begin{enumerate}
\vspace{-0mm}
\item Compute a preliminary GI map using Eq.~\ref{eq:grayness_newgreypixel}.
\vspace{-0mm}
\item Discard pixels in $GI$ with no spatial cues using Eq.~\ref{eq:delta_rgbneq0}. To weaken the effect of isolated gray pixels mainly due to camera noise, $GI$ map is averaged in $7\times 7$ window.
\end{enumerate}
For illustration, Fig. \ref{fig:flowchart} shows a flowchart of computing GI and its predicted illumination.

The proposed GI differs from GP in two important aspects. At first,
it utilizes a novel mechanism to detect gray pixels based on a more complete image formation model that leads to different formulation. Secondly, the proposed GI
works without selectively enhancing bright and dark pixels according to their luminance. In other words, the proposed GI does not weaken the influence of dark pixels.

\subsection{GI Application in Color Constancy}
Color Constancy is a direct application of gray pixels. Here we describe two pipelines to compute illumination color from gray pixels: {\em single illumination} and
{\em multi-illumination} pipelines.

When a scene contains only one global illumination, the pipeline is straightforward. As shown in Fig.~\ref{fig:flowchart}, after ranking all image pixels according to their GI, the global illumination is computed as the average of top $N\%$ pixels. 

Given a scene cast by more than one light source, the desired output is a pixel-wise illumination map.  Similar to~\cite{yang2015efficient}, the GI map is first computed and then followed by a K-means clustering of the top $N\%$ pixels into
preset number of $M$ clusters. Now, the averaging is applied on cluster basis, giving a illumination vector $L_m$ for the cluster $m$. The final spatial illumination map is 
computed using: 

\vspace{-5mm}
\begin{equation}
\begin{split}
\label{eq:kmean}
L_i(x,y)=\sum_{m=1}^{M} \omega_m L^i_m, i \in \{R,G,B\}
\end{split}
\end{equation}
where $\omega_m$ controls the connection between the pixel $I(x,y)$ to the cluster $m$, written as:

\vspace{-5mm}
\begin{equation}
\begin{split}
\label{eq:d_kmean}
\omega_m=e^{-\frac{D_m}{2\sigma^2}} / \sum_{n=1}^{M} e^{-\frac{D_n}{2\sigma^2}}, 
\end{split}
\end{equation}
where $D_m$ is the Euclidean distance from the pixel to the centroid of cluster $m$. Eq.~\ref{eq:d_kmean} encourages nearby pixels to share a similar illumination.

\section{Evaluation}
\label{sec:experiments}


We evaluated GI in two color constancy settings: (1) single-illumination estimation, where the illumination of the whole captured scene is described by a single chroma vector for the red, green and blue channels; and (2) multi-illumination estimation, where in each scene there are two or more effective illuminants. Moreover, we conducted experiments in the cross-dataset setting which is very challenging for the learning-based methods.  


\paragraph{Datasets}
\begin{compactitem}
\item The Gehler-Shi Dataset \cite{shi2010re,gehler2008bayesian}: single illumination, $568$ high dynamic linear images, $2$ cameras \footnote{cameras: Canon 1D, Canon 5D}.
\item The NUS 8-Camera Dataset \cite{Cheng14}: single illumination, $1,736$ high dynamic linear images, $8$ cameras (see Table~\ref{tab:nus_camera_invariance} for the camera list).
\item MIMO Dataset \cite{beigpour2014multi}: multi-illumination, $78$ linear images, $58$ laboratory images and $20$ harder wild images.
\end{compactitem}

\paragraph{Single-illumination Experiment Settings}
\begin{compactitem}
\item The local contrast operator in Eq.~\ref{eq:grayness_newgreypixel} is the Laplacian of Gaussian filter of the size $5$ pixels. 
\item  The proportion of the best gray pixels used for color estimation is set to $N=0.1\%$.
\item The contrast threshold is set to $\epsilon=1\mathrm{e}{-4}$
\end{compactitem}
These parameters were selected based on preliminary grid search (see Section~\ref{sec:params}) and remained fixed for all experiments with the both datasets.

\paragraph{Multi-illumination Experiment Settings}
\begin{compactitem}
\item The local contrast operator and the contrast threshold are the same as in the single-illuminant experiment.
\item  The proportion of chosen pixels is set to $N=10.0\%$ as more illuminants are involved.
\item  The tested number of clusters $M$ were $2$,$4$ and $6$.
\end{compactitem}

\paragraph{Dataset Bias of Learning-based Methods}
When trained with images from a single data that is divided to training and testing sets, the state-of-the-art learning-based methods (\eg \cite{barron2017fourier}) outperform the best learning-free methods by a clear margin. However, it is important to know how these values are biased since images in the training and test sets often share the same camera(s) and same scenes. It can happen that a learning-based method overfits to the camera and
scene features that are not available in the real case. To investigate the dataset bias, we evaluated several top performing learning-based methods in the cross-dataset setting, where the methods were trained on one dataset (\textit{e.g.}, the Gehler-Shi) and tested with another. This allows evaluating the performance of learning-based algorithms for unseen cameras and scenes.


\paragraph{Performance Metric}
As the standard tool in color constancy papers we adopted
the angular error $\arccos(\frac{L^T\hat{L}}{\Vert L \Vert \Vert \hat{L} \Vert})$ between the estimated illumination $\hat{L}$ and ground-truth $L$ as the performance metric. Obtained results are summarized in Table~\ref{tab:maintable} and discussed in Sections~\ref{sec:known} and \ref{sec:agnostic}.

\begin{table*}
\caption{Quantitative Evaluation of CC methods. All values correspond to angular error in degrees. We report the results of the related work in the following order: 1) the cited paper, 2) Table [1] and Table [2] from Barron \textit{et al.}~\cite{barron2017fourier,barron2015convolutional} considered to be up-to-date and comprehensive, 3) the color constancy benchmarking website~\cite{colorconstancy.com}. We left dash on unreported results. In (a) results of learning-based methods worse than ours are marked in gray. The training time and testing time are reported in seconds, averagely per image, if reported in the original paper.   
}
\label{tab:maintable}

\begin{threeparttable}
\centering
\subfloat[single-dataset setting]{
\label{tab:known}
\resizebox{1.0\linewidth}{!}{
\begin{tabular}{l  rr rrr || rr rrr }

\toprule
    & \multicolumn{5}{c}{Gehler-Shi} & 
    \multicolumn{5}{c}{{NUS 8-camera}}\\    
    
  & { Mean} & { Median} & Trimean & Best 25\% & Worst 25\% &{ Mean} & { Median} & Trimean & Best 25\% & Worst 25\% \\
\midrule
\multicolumn{5}{c}{{\em Learning-based Methods (camera-known setting)}} \\
Edge-based Gamut ~\cite{gijsenij2010generalized}
 & \cellcolor{gray!25}6.52 & \cellcolor{gray!25}5.04 & \cellcolor{gray!25}5.43 & \cellcolor{gray!25}1.90 & \cellcolor{gray!25}13.58&  \cellcolor{gray!25}4.40  & \cellcolor{gray!25}3.30  & \cellcolor{gray!25}3.45 & \cellcolor{gray!25}0.99 & \cellcolor{gray!25}9.83\\
 Pixel-based Gamut~\cite{gijsenij2010generalized}
 & \cellcolor{gray!25}4.20 & \cellcolor{gray!25}2.33 & \cellcolor{gray!25}2.91 & \cellcolor{gray!25}0.50 & \cellcolor{gray!25}10.72 &  \cellcolor{gray!25}5.27  & \cellcolor{gray!25}4.26  & \cellcolor{gray!25}4.45 & \cellcolor{gray!25}1.28 & \cellcolor{gray!25}11.16 \\
Bayesian \cite{gehler2008bayesian}
& \cellcolor{gray!25}4.82 & \cellcolor{gray!25}3.46  &  \cellcolor{gray!25}3.88 & \cellcolor{gray!25}1.26 & \cellcolor{gray!25}10.49 & \cellcolor{gray!25}3.50  & \cellcolor{gray!25}2.36 & \cellcolor{gray!25}2.57 & \cellcolor{gray!25}0.78 & \cellcolor{gray!25}8.02\\
Natural Image Statistics~\cite{gijsenij2011color}
  & \cellcolor{gray!25}4.19 & \cellcolor{gray!25}3.13 & \cellcolor{gray!25}3.45 & \cellcolor{gray!25}1.00 & \cellcolor{gray!25}9.22& \cellcolor{gray!25}3.45 &  \cellcolor{gray!25}2.88  & \cellcolor{gray!25}2.95 &\cellcolor{gray!25}0.83 & \cellcolor{gray!25}7.18 \\
  
 Spatio-spectral (GenPrior) \cite{chakrabarti2012color}
 & \cellcolor{gray!25}3.59  & \cellcolor{gray!25}2.96& \cellcolor{gray!25}3.10 & \cellcolor{gray!25}0.95 & 7.61 &\cellcolor{gray!25}3.06  & \cellcolor{gray!25}2.58  & \cellcolor{gray!25}2.74 & \cellcolor{gray!25}0.87 & 6.17 \\

Corrected-Moment\tnote{1} (19 Edge)  \cite{finlayson2013corrected}
&  \cellcolor{gray!25}3.12 & \cellcolor{gray!25}2.38  & \cellcolor{gray!25}2.59   & \cellcolor{gray!25}0.90 & 6.46 
& \cellcolor{gray!25}3.03 & \cellcolor{gray!25}2.11  & \cellcolor{gray!25}2.25 & \cellcolor{gray!25}0.68 & \cellcolor{gray!25}7.08 \\

Corrected-Moment\tnote{1}(19 Color)  \cite{finlayson2013corrected}
&  2.96 & \cellcolor{gray!25}2.15  & \cellcolor{gray!25}2.37   & \cellcolor{gray!25}0.64 & 6.69 
& \cellcolor{gray!25}3.05 & 1.90  & 2.13 & \cellcolor{gray!25}0.65 & \cellcolor{gray!25}7.41 \\

Exemplar-based~\cite{joze2014exemplar}$^*$
  & 2.89 &  \cellcolor{gray!25}2.27  &\cellcolor{gray!25}2.42  & \cellcolor{gray!25}0.82 & 5.97 & --&-- &--& -- &-- \\
  
Chakrabarti \textit{et al.} 2015 ~\cite{chakrabarti2015color}
  & 2.56 &  1.67  & 1.89  &\cellcolor{gray!25} 0.52 & 6.07 & --&-- &--& -- &-- \\
  
Cheng \textit{et al.} 2015 \cite{cheng2015effective}
  & 2.42 &1.65 &1.75 & 0.38 & 5.87 &  2.18  & 1.48  & 1.64 & 0.46 & 5.03 \\
  
DS-Net (HypNet+SelNet) \cite{shi2016eccv}
  & 1.90  & 1.12 & 1.33 & 0.31 & 4.84 &  2.24  & 1.46 & 1.68 & 0.48 & 6.08 \\
  
CCC (dist+ext)  \cite{barron2015convolutional} 
  & 1.95 & 1.22 & 1.38 & 0.35 & 4.76 &  2.38 & 1.48 & 1.69 & 0.45 & 5.85 \\
FC$^{4}$  (AlexNet) \cite{hu2017cvpr}
& \textbf{1.77} & 1.11 & 1.29 & 0.34 & \textbf{4.29} &  2.12 & 1.53 & 1.67 & 0.48 & 4.78\\
FFCC  \cite{barron2017fourier}
  & 1.78 & \textbf{0.96} & \textbf{1.14} & \textbf{0.29} & 4.62 &  \textbf{1.99} & \textbf{1.31} & \textbf{1.43} & \textbf{0.35} & \textbf{4.75}\\
\midrule

\textbf{GI}
& {3.07} &  {1.87} & {2.16} & {0.43} & 7.62
& {2.91} & {1.97} & {2.13} & {0.56} & 6.67\\
\bottomrule
\end{tabular}
}
}

\begin{flushleft}
\scriptsize
$^1$ For Correct-Moment \cite{finlayson2013corrected} we report reproduced and more detailed results by \cite{barron2015convolutional}, which slightly differs with the original results: {mean: 3.5, median: 2.6} for 19 colors and {mean: 2.8, median: 2.0} for 19 edges on Gehler-Shi Dataset.\\
$^*$ We mark Exemplar-based method with asterisk as it is trained and tested on a uncorrected-blacklevel dataset. 
\end{flushleft}

\end{threeparttable}

\begin{center}
\begin{threeparttable}
\subfloat[cross-dataset setting]{
\label{tab:agnostic}

\resizebox{\linewidth}{!}{
\begin{tabular}{l  rr rrr || rr rrr || rr }
\toprule

Training set	&  \multicolumn{5}{c}{{NUS 8-Camera}}  & \multicolumn{5}{c}{{Gehler-Shi}}&\multicolumn{2}{c}{Average}\\
Testing set 	&  \multicolumn{5}{c}{{Gehler-Shi}}  & \multicolumn{5}{c}{{NUS 8-Camera}}&\multicolumn{2}{c}{runtime (s)}\\
    
				& { Mean} & { Median} & Trimean & Best 25\% & Worst 25\% &{ Mean} & { Median} & Trimean & Best 25\% & Worst 25\% & Train & Test\\

\midrule

\multicolumn{5}{c}{{\em Learning-based Methods (agnostic-camera setting), Our rerun}} \\
Bayesian \cite{gehler2008bayesian} & 4.75 & 3.11 & 3.50 & 1.04 & 11.28 & 3.65 & 3.08 & 3.16 & 1.03 & 7.33 & 764 & 97\\
Chakrabarti \textit{et al.} 2015 ~\cite{chakrabarti2015color} Empirical&3.49 & 2.87 & 2.95 & 0.94 & {\bf 7.24} & 3.87 & 3.25 & 3.37 & 1.34 & 7.50 & -- & 0.30\\
Chakrabarti \textit{et al.} 2015 ~\cite{chakrabarti2015color} End2End&  3.52 & 2.71 & 2.80 & 0.86 & 7.72 & 3.89 & 3.10 & 3.26 & 1.17 & 7.95 & -- & 0.30\\

Cheng \textit{et al.} 2015 \cite{Chen-2015-CVPR}  &  5.52 &   4.52 & 4.79& 1.96 & 12.10   & 4.86  & 4.40  &  4.43& 1.72 & 8.87 & 245 & 0.25\\ 
FFCC \cite{barron2017fourier}  & 3.91 &  3.15 & 3.34 & 1.22 & {7.94} & 3.19 &  2.33 & 2.52 & 0.84 & 7.01 & 98 & 0.029\\ 
\midrule
\multicolumn{5}{c}{{\em Physics-based Methods}} \\
IIC~\cite{tan2008color} & 13.62 &  13.56 & 13.45  & 9.46 & 17.98 & -- & -- & -- & -- & -- & -- &  --\\
Woo \textit{et al.} 2018~\cite{sung2018tip} & 4.30 &  2.86 & 3.31  & 0.71 & 10.14 & -- & -- & -- & -- & -- & -- &  --\\
\midrule
\multicolumn{5}{c}{{\em Biological Methods}} \\
Double-Opponency~\cite{gao2015color} & 4.00 &  2.60 & --  & -- & -- & -- & -- & -- & -- & -- & -- &  --\\
ASM 2017~\cite{akbarinia2017colour} & 3.80 &  2.40 & 2.70  & -- & -- & -- & -- & -- & -- & -- & -- &  --\\
\midrule
\multicolumn{5}{c}{{\em Learning-free Methods}} \\
White Patch \cite{brainard1986analysis}
& 7.55 &  5.68 & 6.35 & 1.45 & 16.12 & 9.91  & 7.44 &  8.78  & 1.44 & 21.27  & -- & 0.16\\
Grey World~\cite{buchsbaum1980spatial} & 6.36 &  6.28 & 6.28  & 2.33 & 10.58 & 4.59  & 3.46 &  3.81 & 1.16 & 9.85 & -- & 0.15\\
General GW~\cite{barnard2002comparison}
& 4.66  & 3.48 & 3.81 & 1.00 & 10.09 & 3.20 &  2.56 & 2.68 & 0.85 & 6.68& -- & 0.91\\ 
 2st-order grey-Edge ~\cite{van2007edge}
& 5.13 &4.44& 4.62 & 2.11 & 9.26  & 3.36  & 2.70  &2.80 & 0.89 & 7.14& -- & 1.30\\
1st-order grey-Edge ~\cite{van2007edge}

 & 5.33&4.52& 4.73 & 1.86 & 10.43 &  3.35  & 2.58 & 2.76 & 0.79 & 7.18& -- & 1.10\\
Shades-of-grey \cite{finlayson2004shades}
& 4.93 & 4.01 & 4.23 & 1.14 & 10.20 &  3.67  & 2.94 & 3.03 & 0.99 & 7.75& -- & 0.47\\

Grey Pixel (edge)~\cite{yang2015efficient} & 4.60 &  3.10 & --  & -- & -- & 3.15 & 2.20 & -- & -- & -- & -- &  0.88\\

LSRS  \cite{gao2014eccv}
&  3.31 & 2.80  & 2.87   & 1.14 & 6.39 
& 3.45 & 2.51  & 2.70 & 0.98 & 7.32 & -- & 2.60\\

Cheng \textit{et al.} 2014 \cite{Cheng14}
& 3.52 & 2.14& 2.47 & 0.50 & 8.74 & 2.93 & 2.33  & 2.42  & 0.78 & \textbf{ 6.13}& -- & 0.24\\

\textbf{GI}
& \textbf{3.07} &  \textbf{1.87} & \textbf{2.16} & \textbf{0.43} & 7.62
& \textbf{2.91} & \textbf{1.97} & \textbf{2.13} & \textbf{0.56} & {6.67}& -- & 0.40\\
\bottomrule
\end{tabular} }
}
\end{threeparttable}
\end{center}

\end{table*}

\begin{table*}
\begin{center}
\caption{Each-camera evaluation on the NUS 8-Camera Dataset. \textit{Std} in the last column refers to the standard deviation of statistics (\textit{e.g.} mean angular error) on 8 cameras.} 
\label{tab:nus_camera_invariance}
\resizebox{1.\linewidth}{!}{
  \begin{tabular}{l cccc cccc || c}
    \toprule 
  &  \multicolumn{9}{c}{{NUS 8-camera Dataset}} \\
  & Canon & Canon & Fujifilm & Nikon& Olympus & Panasonic & Samsung & Sony  &  ~~~Std~\\
    & 1DS Mark3 & 600D & X-M1 & D5200 & E-PL6 & DMC-GX1 & NX2000 & SLT-A57  &  ~~~~\\
 \midrule
 \multicolumn{1}{c}{{\em Cheng et al. 2014 \cite{Cheng14}}} \\
Mean  & 2.93 & 2.81  &  3.15 & 2.90 & 2.76 & 2.96  &  2.91 & 2.93  & 0.1152 \\ 
Median  & 2.01 & 1.89  &  2.15 & 2.08 & 1.87 & 2.02  & 2.03 & 2.33  & 0.1465 \\ 
Tri  & 2.22 & 2.12  &  2.41 & 2.19 & 2.05 & 2.31  &  2.22 & 2.42 & 0.1309   \\ 
Best-$25\%$  & 0.59 & 0.55  &  0.65 & 0.56 & 0.55 & 0.67  &  0.66 & 0.78  & 0.0798 \\ 
Worst-$25\%$  & 6.82 & 6.50  &  7.30 & 6.73 & 6.31 & 6.66 &  6.48 & 6.13  & 0.3558\\ 
\midrule
 \multicolumn{5}{c}{{\em Chakrabarti \etal~\cite{chakrabarti2015color} (best), trained on Gehler-Shi, tested here}} \\
Mean    & 3.00 & 3.26  &  3.12 & 3.26 & 3.31 & 3.30  & 3.30 & 3.32 & {0.1056}  \\ 
Median  & 2.17 & 2.48  &  2.45 & 2.48 & 2.50 & 2.49  &  2.48 & 2.56  & {0.1171} \\ 
Tri     & 2.31 & 2.64  &  2.60 & 2.64 & 2.72 & 2.69  &  2.68 & 2.75  & {0.1365} \\ 
Best-$25\%$  & 0.74 & 0.83  & 0.83 & 0.83 & 0.85 & 0.84  &  0.83 & 0.86  & {0.0390} \\ 
Worst-$25\%$  & 6.77 & 7.04  &  6.89 & 7.04 & 7.11 & 7.12  &  7.16 & 7.12 & \textbf{0.1312}  \\ 
\midrule

 \multicolumn{1}{c}{{\em GI}} \\
Mean  & 3.02 & 2.85  &  2.89 & 2.85 & 2.84 & 2.86  &  2.86 & 2.75 & \textbf{0.0753}  \\ 
Median  & 1.87 & 1.96  &  1.98 & 1.96 & 1.97 & 1.97  &  1.97 & 1.89  & \textbf{0.0420} \\ 
Tri  & 2.16 & 2.12  &  2.15 & 2.12 & 2.15 & 2.17  &  2.13 & 2.07  & \textbf{0.0321} \\ 
Best-$25\%$  & 0.54 & 0.55  &  0.55 & 0.55 & 0.56 & 0.56  &  0.55 & 0.53  & \textbf{0.0114} \\ 
Worst-$25\%$  & 7.29 & 6.79  &  6.86 & 6.79 & 6.70 & 6.75  &  6.81 & 6.51 & {0.2198}  \\ 
\bottomrule
  \end{tabular}
  }
\end{center}\vspace{-3mm}
\end{table*}

\begin{table}
\scriptsize
\begin{center}
\caption{Quantitative Evaluation on the MIMO dataset.} 
\label{tab:mimo}
\resizebox{1.\linewidth}{!}{
  \begin{tabular}{l cccc}
    \toprule 
  &  \multicolumn{2}{c}{Laboratory(58)} &  \multicolumn{2}{c}{Real-world(20)} \\
 Method & Median & Mean & Median & Mean \\
 \midrule 
 Doing Nothing & 10.5 & 10.6 & 8.8 & 8.9 \\
Gijsenij \etal \cite{gijsenij2012improving} & 4.2 & 4.8 & 3.8 & 4.2 \\
CRF \cite{beigpour2014multi} & 2.6 & 2.6 & 3.3 & 4.1 \\
GP (best) \cite{yang2015efficient} & 2.20 & 2.88 & 3.51 & 5.68 \\
\midrule
GI (M=2) & 2.09 & 2.66 & \textbf{3.32} & \textbf{3.79} \\
GI (M=4) & 2.09 & 2.65 & 3.47 & 3.96 \\
GI (M=6) & \textbf{2.07} & \textbf{2.60} & 3.49 & 3.94 \\
\bottomrule
  \end{tabular}
  }
\end{center}\vspace{-3mm}
\end{table}


\setlength{\tabcolsep}{1pt}
\renewcommand{\arraystretch}{1}
\begin{figure}[t]
\begin{center}
\hspace{-3ex}
\begin{tabular}{c c}
\raisebox{1\height}{\rotatebox{90}{\scriptsize{{0.77}}}}
&{{\includegraphics[width=\linewidth]{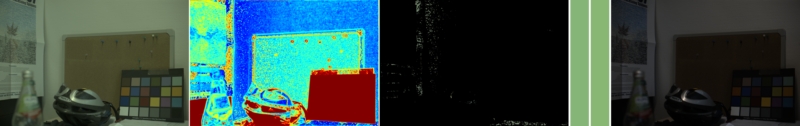}}} \\
\vspace{-0.75mm}
\raisebox{1\height}{\rotatebox{90}{\scriptsize{{0.81}}}}
&{{\includegraphics[width=\linewidth]{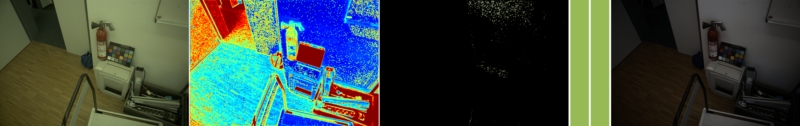}}} \\
\vspace{-0.75mm}
\raisebox{1\height}{\rotatebox{90}{\scriptsize{{2.78}}}}
&{{\includegraphics[width=\linewidth]{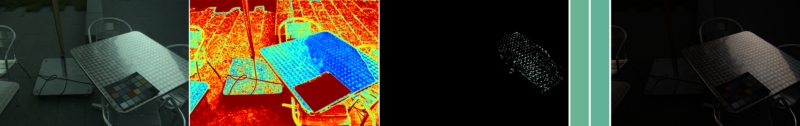}}} \\
\vspace{-0.75mm}
\raisebox{1\height}{\rotatebox{90}{\scriptsize{{2.50}}}}
&{{\includegraphics[width=\linewidth]{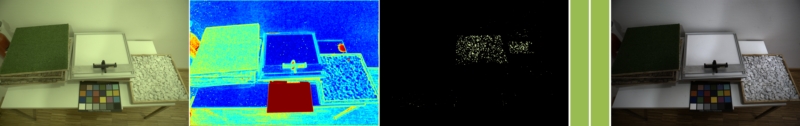}}} \\

\end{tabular}

\end{center}
\caption{Qualitative results on the single-illumination Gehler-Shi. From left to right: angular error, input image, GI, top $1\%$ pixels chosen as gray pixel, estimated illumination color, the ground truth color and corrected image using the predicted illumination.
Macbeth Color Checker is always masked as GI finds perfect gray patch as gray pixels.
}
\vspace{-5mm}
\label{fig:single-illu}
\end{figure}

\subsection{Single-dataset Setting}
\label{sec:known} 
Single-dataset setting is the most common setting in related works, allowing extensive pre-training using k-fold cross-validation for learning-based methods. The results for this setting are summarized in Table~\ref{tab:known}. Among all the compared methods, up to the date of submission of this paper, FFCC~\cite{barron2017fourier} achieves the best overall performance with the both datasets. It is important to remark that cross-validation makes no difference to the performance of statistical methods. Therefore, in order to avoid repetition, the performance of competing non-learning methods are shown only once in Table \ref{tab:agnostic}. For visualization purposes, results of learning-based methods that are outperformed by the proposed GI are highlighted in gray. Remarkably, it is clear that, even in the setting which is friendly to learning-based method, GI outperforms several popular learning-based methods (from Gamut~\cite{gijsenij2010generalized} to the industry-standard Corrected-Moment~\cite{finlayson2013corrected}) without the need of extensive training and parameter tuning. Visual examples of GI are shown in Fig. \ref{fig:single-illu}.

Comparing to the best learning-based methods (\textit{e.g.} \cite{chakrabarti2015color}), GI has a noticeable heavy tail in its angular error distribution ({\em e.g.} amont the worst $25\%$ cases), which suggests that GI would be more optimal if gray pixels would be i.i.d over the whole datasets (\textit{e.g.} natural images). Learning-based methods perform well on these ``rare'' cases using 3-fold cross-validation, and can further improve ``rarity case'' performance by including more training data (\textit{e.g.} via 10-fold cross-validation)~\cite{chakrabarti2015color}.

\setlength{\tabcolsep}{1pt}
\renewcommand{\arraystretch}{1}
\begin{figure}[t]
\begin{center}

\begin{tabular}{c cccc}

&{{\includegraphics[width=1.8cm]{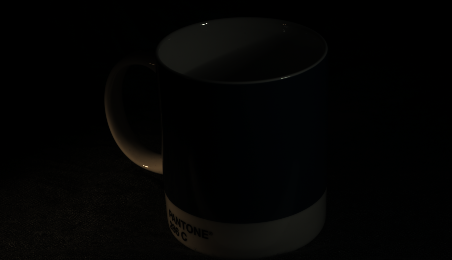}}}
&{{\includegraphics[width=1.8cm]{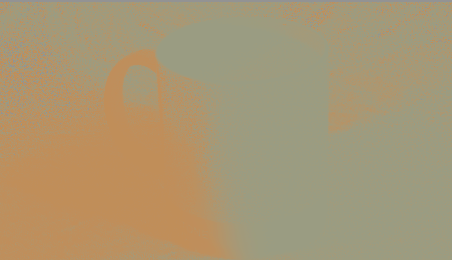}}}
&{{\includegraphics[width=1.8cm]{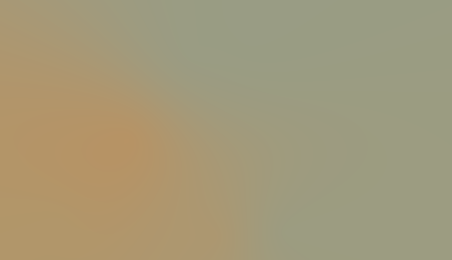}}}
&{{\includegraphics[width=1.8cm]{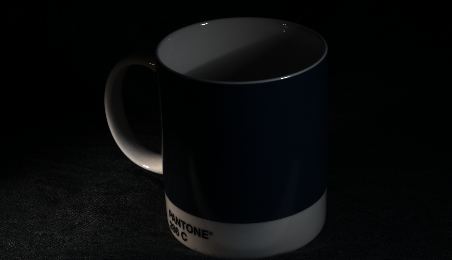}}}
\\
\vspace{-0.75mm}
&{{\includegraphics[width=1.8cm]{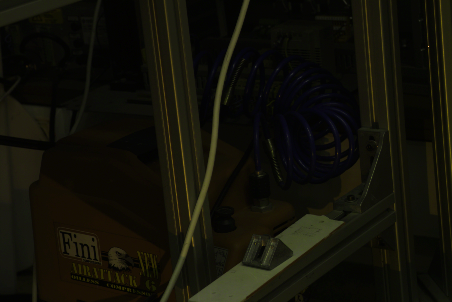}}}
&{{\includegraphics[width=1.8cm]{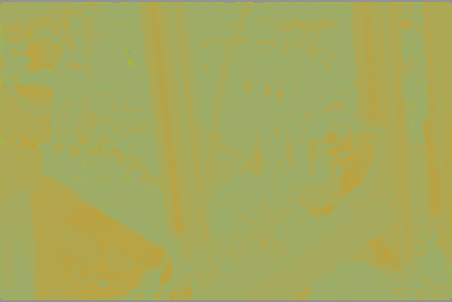}}}
&{{\includegraphics[width=1.8cm]{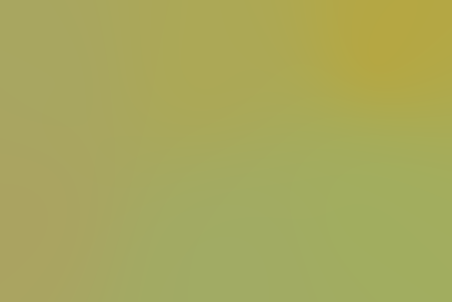}}}
&{{\includegraphics[width=1.8cm]{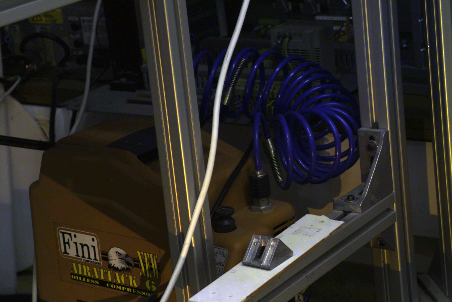}}}
\\
\vspace{-0.75mm}
&{{\includegraphics[width=1.8cm]{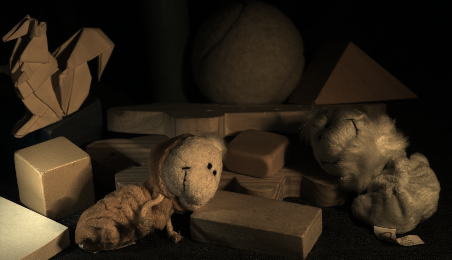}}}
&{{\includegraphics[width=1.8cm]{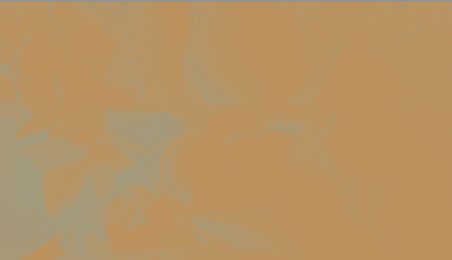}}}
&{{\includegraphics[width=1.8cm]{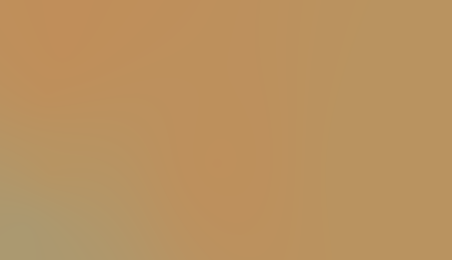}}}
&{{\includegraphics[width=1.8cm]{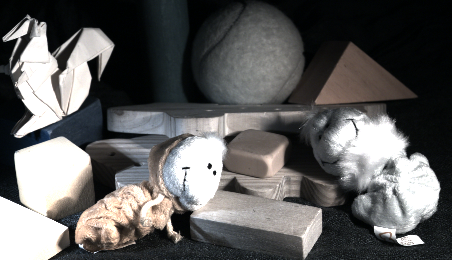}}}
\\

\vspace{-0.75mm}
&{{\includegraphics[width=1.8cm]{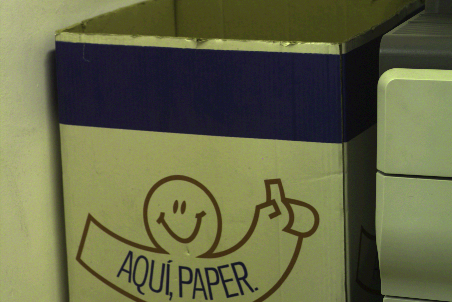}}}
&{{\includegraphics[width=1.8cm]{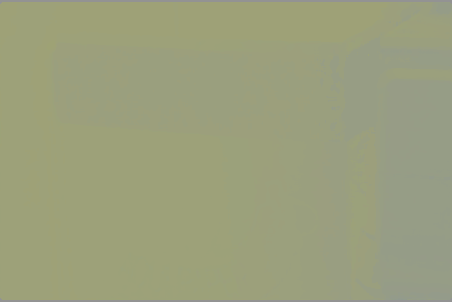}}}
&{{\includegraphics[width=1.8cm]{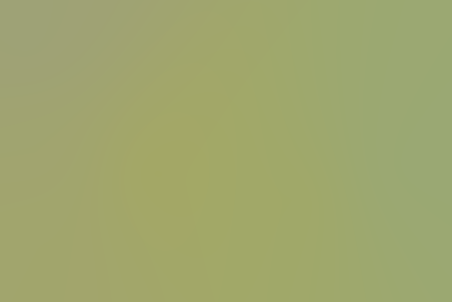}}}
&{{\includegraphics[width=1.8cm]{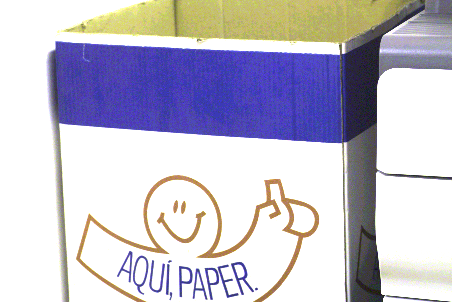}}}
\\
\vspace{-0.75mm}
&{{\includegraphics[width=1.8cm]{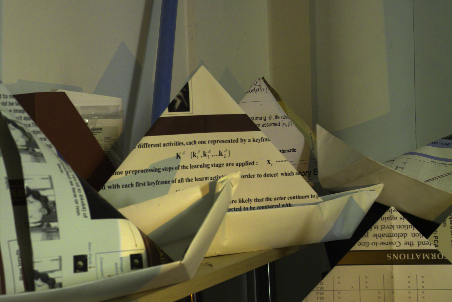}}}
&{{\includegraphics[width=1.8cm]{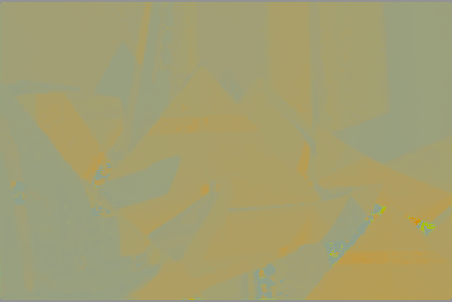}}}
&{{\includegraphics[width=1.8cm]{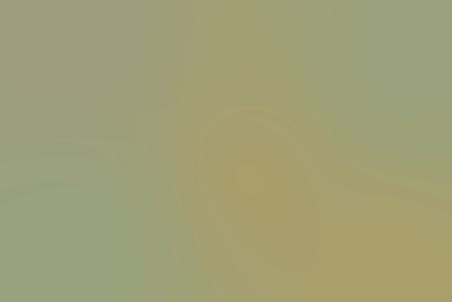}}}
&{{\includegraphics[width=1.8cm]{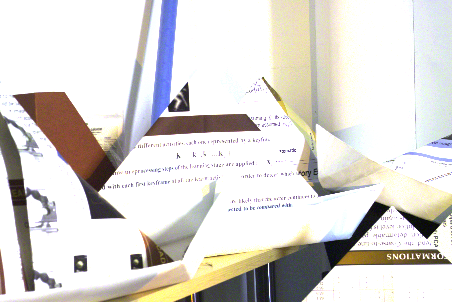}}}

\vspace{-3mm}

\end{tabular}

\end{center}
\vspace{-2.5mm}
\caption{Qualitative results on (multi-illumination) MIMO dataset. From left to right, color-biased input, groundtruth spatial illumination, our spatial estimation using GI, our corrected image. }
\label{figure:mimo}
\vspace{-3mm}
\end{figure}


\subsection{Cross-Dataset Setting}
\label{sec:agnostic}
We were able to re-run the Bayesian method~\cite{gehler2008bayesian}, Chakrabarti~\textit{et al.}\cite{chakrabarti2015color}, FFCC~\cite{barron2017fourier}, and the method by Cheng \textit{et al.} 2015~\cite{cheng2015effective}, using the codes provided by the original authors. Note that
this list of methods includes FFCC, which showed the best overall performance in the camera-known setting. From the provided code we found different approaches to correct the black level and saturated pixels. For consistency, we used a uniform correction process (given in supplement), which was applied to GI as well. 

When we trained on one dataset and tested with another, we made sure that the datasets share no common cameras. For the results reported in this section, we used the best or final setting for each method: Bayes (GT) for Bayesian; Empirical and End-to-End training for Chakrabarti \textit{et al.} ~\cite{chakrabarti2015color}; $30$ regression trees for Cheng \textit{et al.}; full image resolution and $2$ channels for FFCC. Obtained results are summarized in Table~\ref{tab:agnostic}. From this table, it is clear that GI outperforms \textbf{all} learning-based and statistical methods. 

All selected learning-based methods perform worse in this setting, as compared to some statistical methods (\textit{e.g.} LSRS \cite{gao2014eccv}, Cheng \textit{et al.} 2014 \cite{Cheng14}). It is not surprising that the performance of learning-based methods degrades in this scenario. For example, in \cite{barron2017fourier} it is visualized that FFCC models two varying camera sensitivity for Gehler-Shi in preconditioning filter (two wrap-around line segments), which in cross-dataset setting will be improperly used to evaluate performance on the NUS 8-camera Dataset.

\begin{figure}[t]
\begin{center}
\subfloat[]{\includegraphics[width=.27\linewidth]{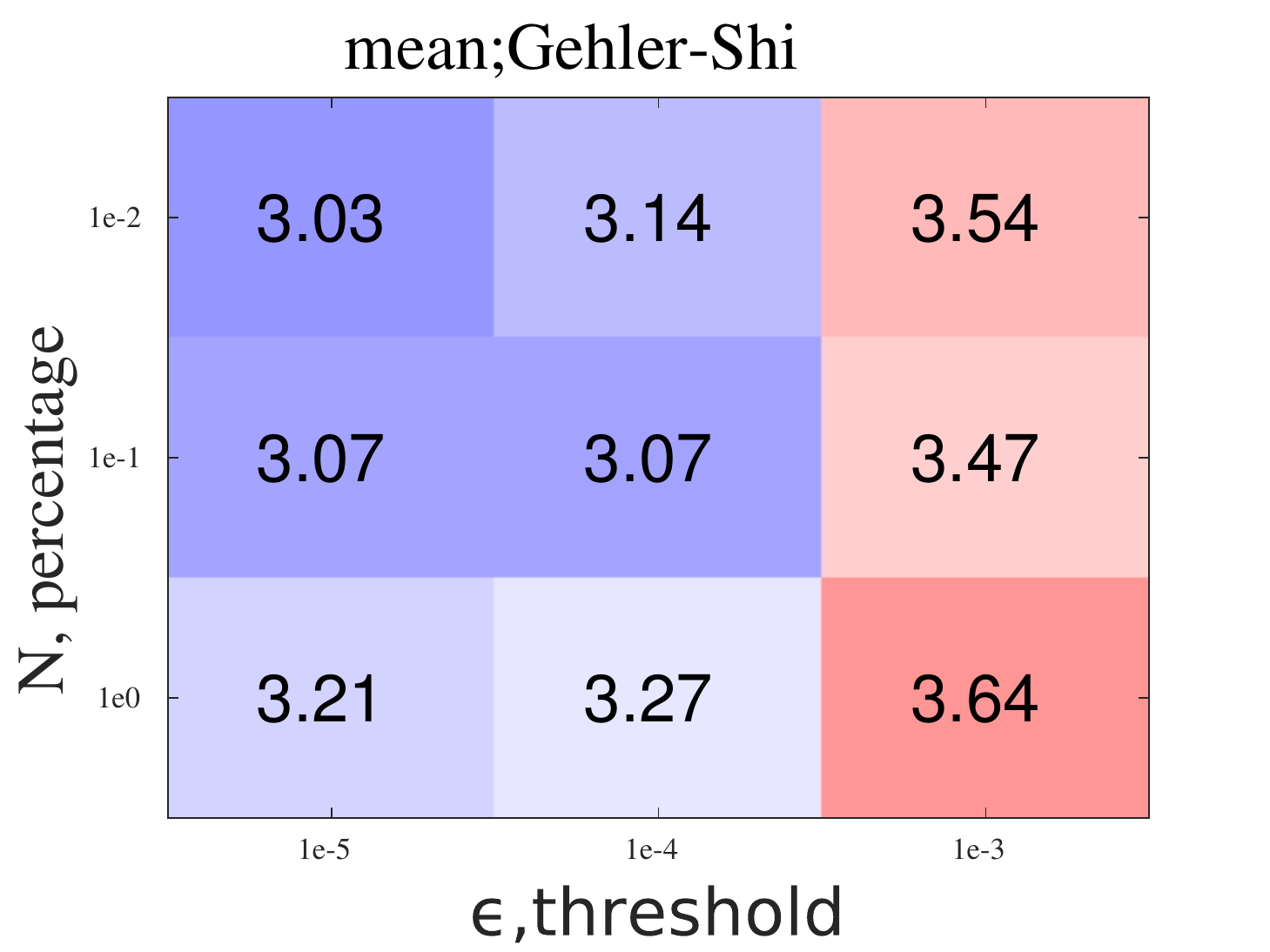}\label{fig:grid_cc_mean}}  
\subfloat[]{\includegraphics[width=.27\linewidth]{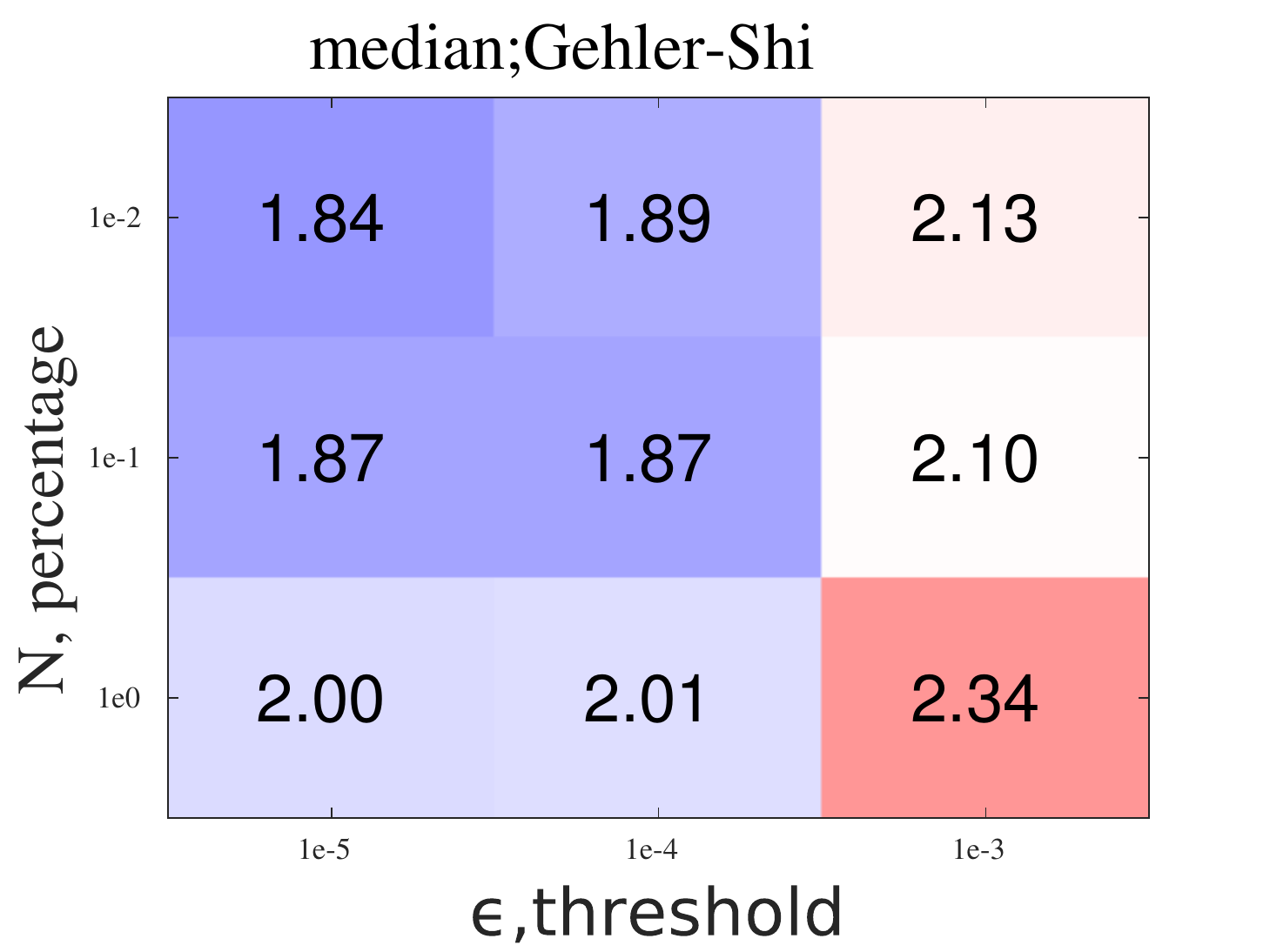}\label{fig:grid_cc_median}}  
\subfloat[]{\includegraphics[width=.27\linewidth]{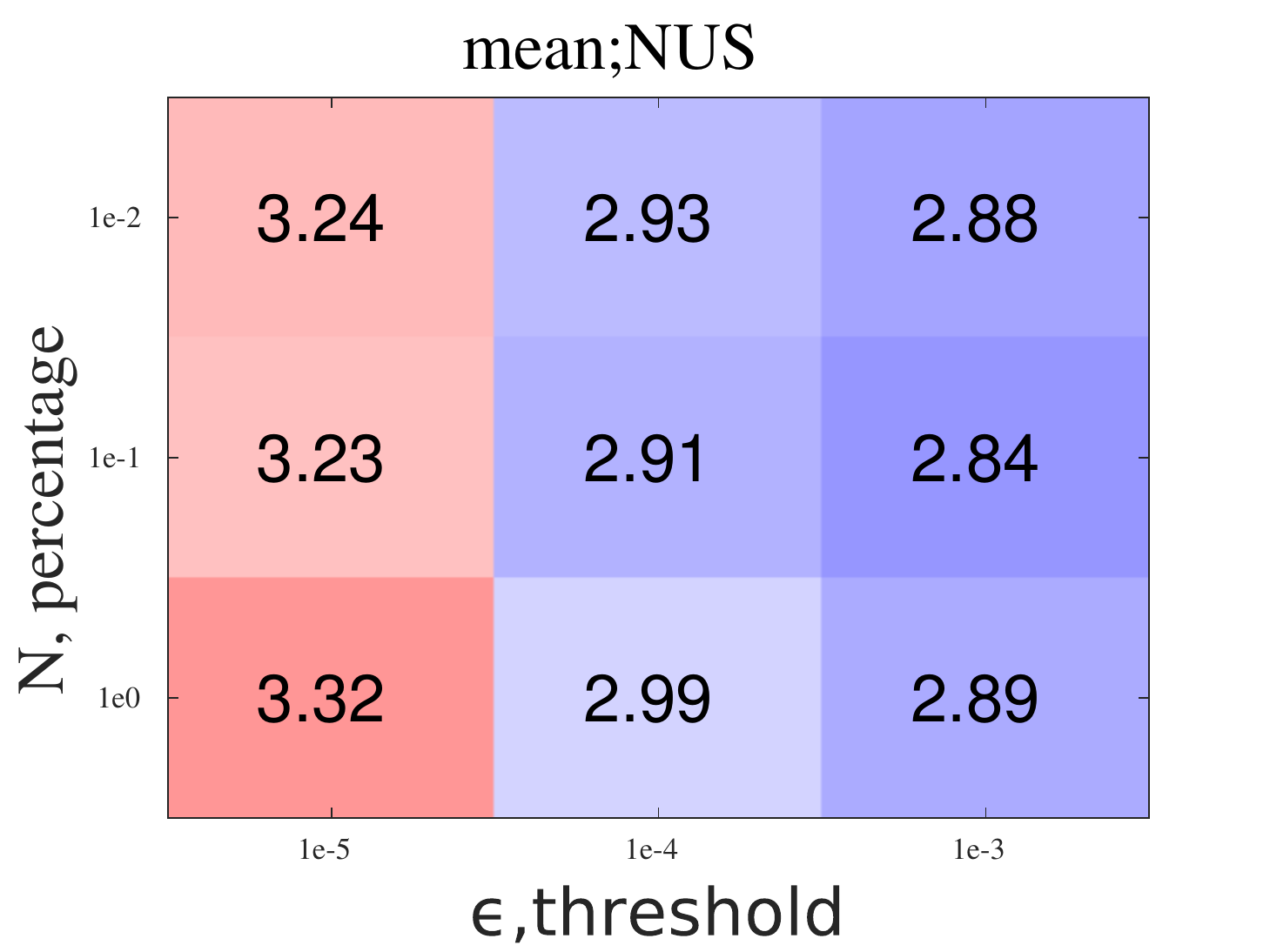}\label{fig:grid_nus_mean}}  
\subfloat[]{\includegraphics[width=.27\linewidth]{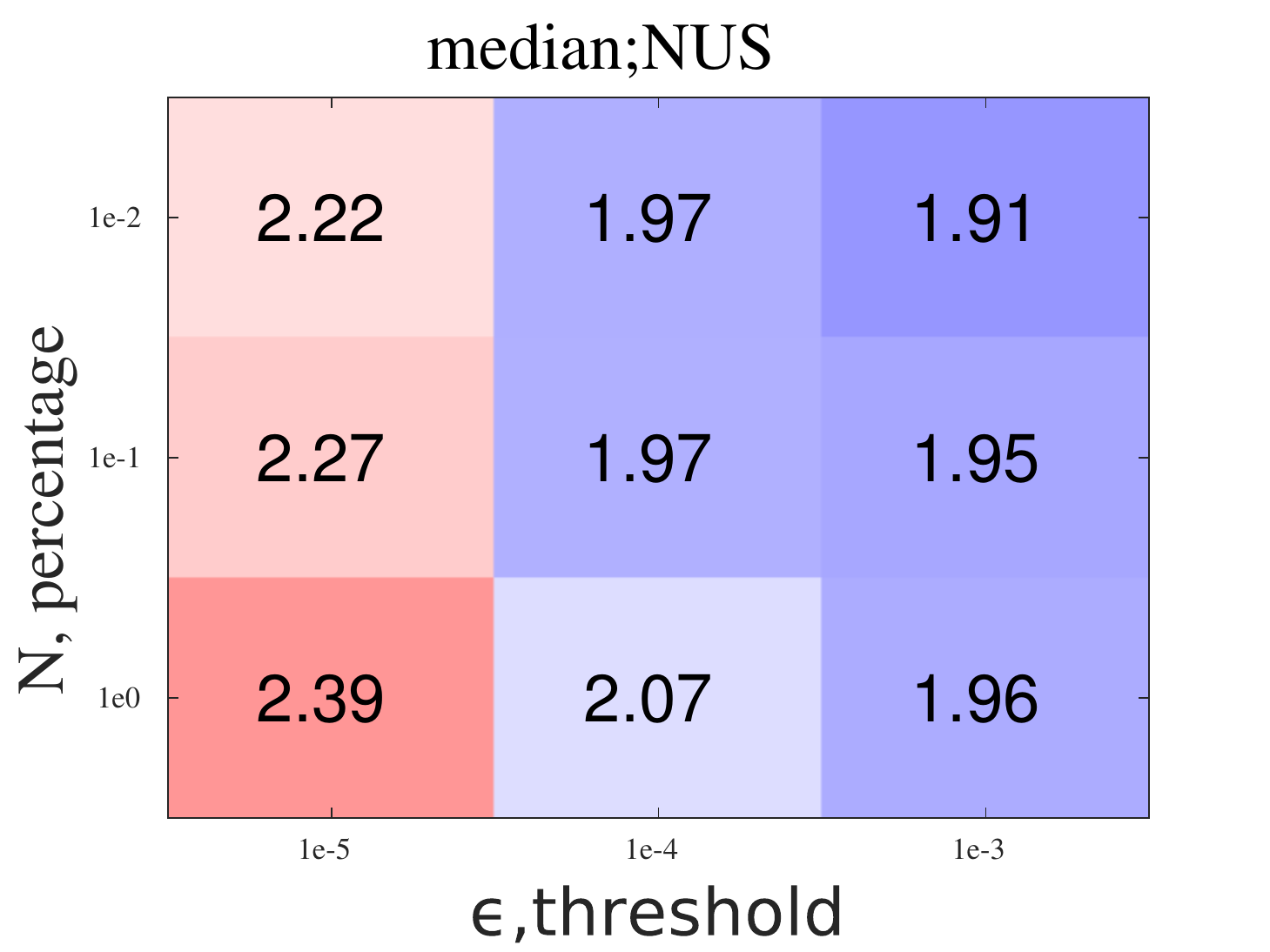}\label{fig:grid_nus_median}}  
\caption{The colormaps of mean and median angular errors corresponding to various $N$ and $\epsilon$ (see the text) for (a,b) Gehler-Shi; (c,d) NUS 8-camera. }
\label{fig:parameter_study}
\end{center}\vspace{-0.8cm}
\end{figure}

A special feature of the NUS 8-Camera Benchmark is that it includes 8 cameras that share the same scenes. We leveraged this feature to evaluate the robustness of the well-performing learning-free and learning-based methods. These results are summarized in Table \ref{tab:nus_camera_invariance}, where GI achieves much more stable results (standard variance is smaller) across 8 cameras. Due to space limitations, we refer readers to \cite{Cheng14} for more results on individual cameras with other methods, including but not limited to \cite{barnard2002comparison,van2007edge,gijsenij2010generalized,gehler2008bayesian,chakrabarti2012color}.
Among all methods in Table 2 of \cite{Cheng14}, GI is less sensitive to camera hardware.

\subsection{Grid Search on Parameters}
\label{sec:params}

The only two parameters in GI are: the percentage $N\%$ of pixels chosen as gray for illumination estimation, and the threshold $\epsilon$ of Eq.~\ref{eq:delta_rgbneq0} used to remove regions without spatial cues. The former restricts the domain range where illumination norm is measured, analogous to the \textit{receptive field} in deep learning, while the later one passes only noticeable activation, like the \textit{ReLU} activation. Figure \ref{fig:parameter_study} summarizes the obtained median and mean angular errors corresponding to a grid search of the parameters, with $N \in \{10^{-2}, 10^{-1}, 1$ and $\epsilon \in \{10^{-5}, 10^{-4}, 10^{-3}\}$, on Gehler-Shi Dataset and NUS 8-camera Dataset. The setting ($N=1\mathrm{e}{-1}$ and $\epsilon=1\mathrm{e}{-4}$) results in a good trade-off between mean and median error on both datasets. The shown parameter grid seems loose, but on the contrary, this shows that our method is robust to parameter tuning across orders of magnitude.

\subsection{Multi-illumination Setting}
\label{sec:multi}
As a side product of grayness index, we evaluate the proposed method on a multi-illumination dataset. Table \ref{tab:mimo} indicates that despite the fact that GI is not designed to deal with spatial illumination changes, it still outperforms well-performing methods \cite{beigpour2014multi, yang2015efficient} with a clear margin. From the mean value over real-world images, it is obvious that GI can better handle
multi-illumination situations. Increasing the number of clusters $M$ from 2 to 6 further improved our results on indoor images, but not for wild ones. Figure \ref{figure:mimo} shows the spatial estimation predicted using GI. Due to Euclidean distance used by the K-means, GI predictions are not sharp in some scenes with complex geometry but still obtain the best overall error rate and plausible visual color correction.

\section{Problems with the ``Ground-truth''}
\label{sec:curious_gt}

\begin{figure}
\centering
\subfloat[]{\includegraphics[width=.35\linewidth]{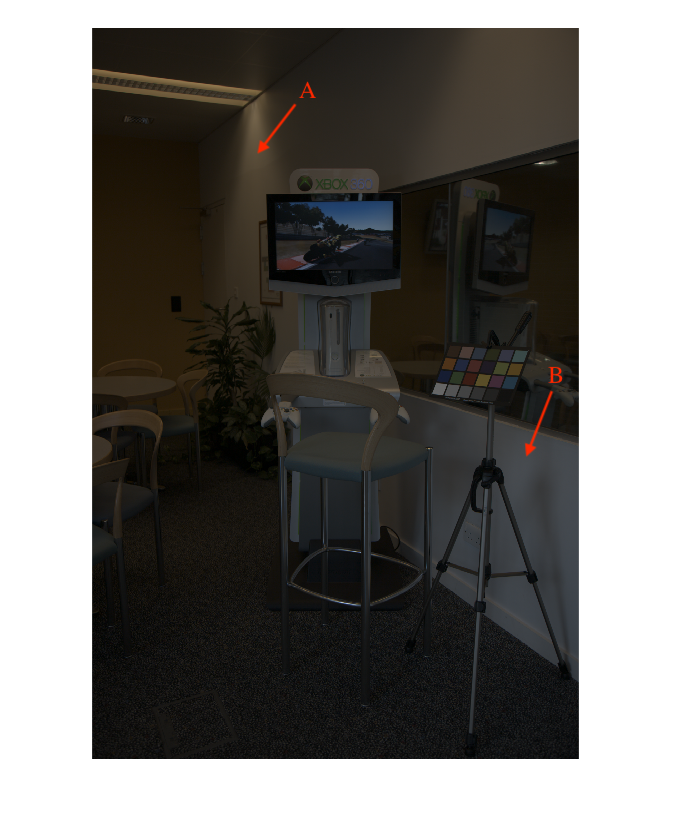}
\label{fig:bbox1}}  
\subfloat[]{\includegraphics[width=.35\linewidth]{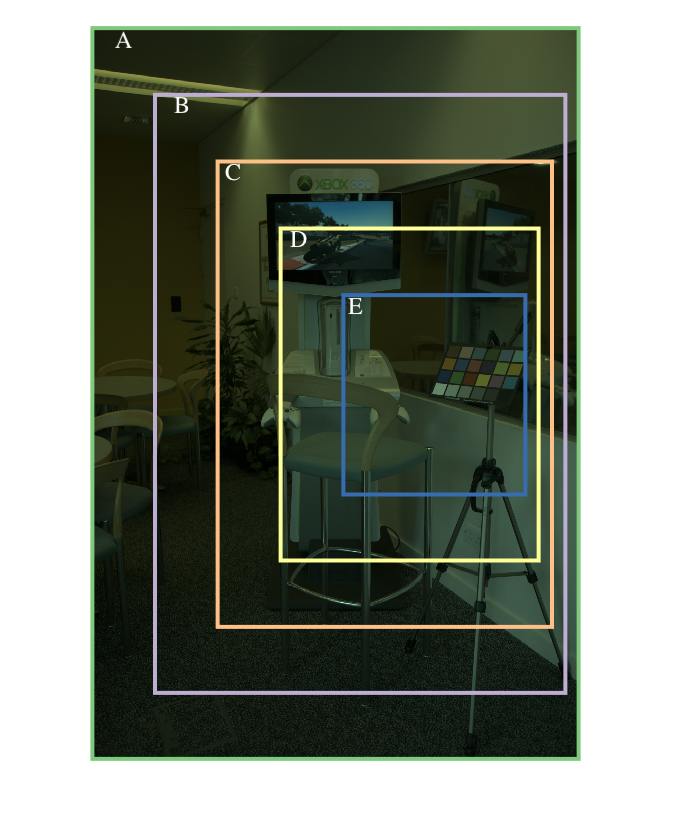}
\label{fig:bbox2}}  
\subfloat[]{\includegraphics[width=.35\linewidth]{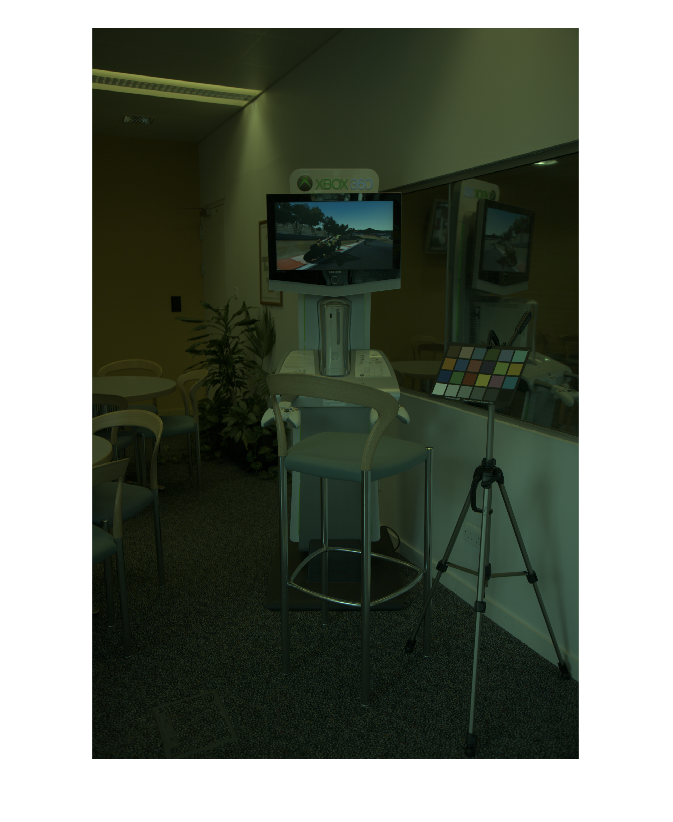}
\label{fig:bbox3}}  
\caption{(a) Example images from the Gehler-Shi corrected using groundtruth, where two different illuminations (red arrow A and B) exist. (b) We test CC methods in decreasing box sizes (from A  to E). (c) Color-biased (a). 
} 
\label{fig:bbox}\vspace{-3mm}
\end{figure}

\begin{table}
\caption{Testing GI, FFCC on varying-size cropped images from Gehler-Shi, given illumination split from \cite{cheng2016cvpr}.} 
\label{tab:twoillu_and_oneillu}
\tiny
\subfloat[Double-illumination Setting]{
\begin{minipage}{.5\linewidth}
\begin{center}
  \label{tab:twoillu}
  \resizebox{1.0\linewidth}{!}{
  \begin{tabular}{l ccccc}

    \toprule 
  &  \multicolumn{5}{c}{{Gehler-Shi: 66 two-illumination Images}} \\
  & Mean & Median & Trimean & Best-25\% & Worst-25\% \\
 \midrule
 \multicolumn{6}{c}{{\em GI}} \\
A  & 6.12 &  4.54 &  5.24  & 0.70 & 13.72 \\ 
B  & 6.06 &  3.88 &  4.90  & 0.92 & 14.08 \\ 
C  & 6.02 &  3.63 &  5.04  & 0.92 & 14.55 \\ 
D  & 5.46 &  3.46 &  4.13  & 0.77 & 13.69 \\ 
E  & \textbf{4.96} &   \textbf{2.94} &   \textbf{3.45}  &  \textbf{0.53} &  \textbf{12.42} \\ 
\midrule
 \multicolumn{6}{c}{{\em FFCC \cite{barron2017fourier}}} \\
A  &  \textbf{3.11} &   \textbf{1.67} &   \textbf{2.25}  &  \textbf{0.44} &  \textbf{8.00} \\ 
B  & 3.44 &  1.84 &  2.39  & 0.42 & 8.69 \\ 
C  & 4.01 &  2.47 &   2.92 & 0.56 & 10.03 \\ 
D  & 4.64 &  3.13 &  3.53  & 0.62 & 11.38\\ 
E  & 4.99 &  3.29 &  3.72  & 0.60 & 11.92 \\ 
\bottomrule
  \end{tabular}
  }
\end{center} 
  \end{minipage}
  }
  \subfloat[Single-illumination Setting]{
\begin{minipage}{.5\linewidth}
\begin{center}
  \label{tab:oneillu}
    \resizebox{1.0\linewidth}{!}{
  \begin{tabular}{l rrrrr}
    \toprule 
  &  \multicolumn{5}{c}{{Gehler-Shi: 502 single-illumination Images}} \\
  & Mean & Median & Trimean & Best-25\% & Worst-25\% \\
 \midrule
 \multicolumn{6}{c}{{\em GI}} \\
A  &  \textbf{2.78} &   \textbf{1.79} &   \textbf{2.03}  &  \textbf{0.41} &  \textbf{6.75} \\ 
B  & 2.95 &  1.86 &  2.12  & 0.41 & 7.28 \\ 
C  & 3.32 &  2.30 &  2.49  & 0.50 & 7.96 \\ 
D  & 3.93 &  2.97 &  3.14  & 0.70 & 8.90 \\ 
E  & 4.81 &  3.79 &  3.94  & 0.82 & 10.74\\ 
\midrule
 \multicolumn{6}{c}{{\em FFCC \cite{barron2017fourier}}} \\
A  &  \textbf{1.68} &   \textbf{0.94} &   \textbf{1.16}  &  \textbf{0.27} &  \textbf{4.22} \\ 
B  & 1.72 &  1.01 &  1.20  & 0.27 & 4.30 \\ 
C  & 1.84 &  1.11 &   1.29 & 0.29 & 4.58 \\ 
D  & 2.13 &  1.29 &  1.43  & 0.36 & 5.45 \\ 
E  & 2.39 &  1.39 &  1.58  & 0.38 & 6.17 \\ 
\bottomrule
  \end{tabular}
  }
\end{center}
  \end{minipage}
  }
\vspace{-0.5cm}
\end{table}

We investigated those cases where GI made erratic predictions (see the supplement for erratic cases) and have observed that, in some images, there exists gray pixels casted by two illumination sources. A similar problem was noticed by Cheng \textit{et al.} \cite{cheng2016cvpr}, who claimed that in the Gehler-Shi \cite{shi2010re}, there are $66$ two-illumination images. An example of this problem is illustrated in Fig. \ref{fig:bbox}. In Fig. \ref{fig:bbox}(a), where pixels near arrows A and B share the same surface (white wall) but have different illuminations, the color of pixel in the neighborhood of B is close to the Macbeth Color Checker (MCC). In such case, our GI does a good job in identifying gray pixels by following the designed rules and finding gray pixels lying in two illuminants, but this comes at the cost of a large angular error. As a first impression, we suppose this is due to the MCC being more dominated by one of the illuminants.  

We designed a simple experiment to investigate our observation. For the list of 66 two-illumination images (given in \cite{cheng2016cvpr}) in the Gehler-Shi and the remaining 502 single-illumination images, we test GI and FFCC~\cite{barron2017fourier} (full resolution, 2 channels, pretrained on whole Gehler-Shi) on images cropped by boxes of decreasing sizes centered at the MCC (from box $A$ to box $E$ in Fig. \ref{fig:bbox2}. Specifically, the boxes are generated by halving the width and height of the
previous box.

The results summarized in Tables \ref{tab:twoillu} and \ref{tab:oneillu} show a crucial fact: in the single-illumination subset, GI yields larger angular errors as the testing box gets smaller (from box $A$ to $E$). In contrast, in the double-illumination subset, this tendency is reversed. It makes sense that the performance of GI decreases as the testing box shrinks since less reference points are available. A reasonable explanation to the abnormal tendency in the two-illumination subset is that the MCC is placed mainly in one illumination, reflecting a biased ``ground-truth''. This problem restricts the upper limit of the performance of GI and possibly also other statistical color constancy methods. Learning-based methods (especially CNN-based method) suffer less from this problem, as they can learn to reason about some structural information, \textit{e.g.} whole-image chroma histogram, the  physical geometry of the scene, the location where MCC is placed. As expected, FFCC performs worse on smaller boxes. Bearing these results in mind, we argue that learning-based methods and statistical methods should be compared by considering their corresponding advantages and limitations in both single-dataset and cross-dataset scenarios. 

\section{Conclusions}
We derived a method to compute grayness in a novel way -- Grayness Index. It relies on the Dichromatic Reflection Model and can detect gray pixels accurately. Experiments performed on the tasks of single-illumination estimation and multi-illumination estimation verified the effectiveness and efficiency of GI. On standard benchmarks, GI estimates illumination more accurately than state-of-the-art learning-free methods in about 0.4 seconds. 
GI has a clear physical interpretation, which we believe can be used for other vision tasks, \eg intrinsic image decomposition. 

Other conclusions also emerged from the research: learning-based methods generally perform worse in the cross-dataset setting; When testing on an image with color checker masked by zeros, learning-based methods can still exploit the location of the color checker and overfit to scene and camera specific features.

\vspace{-2mm}
\section*{Acknowledgments}
\vspace{-3mm}
 This work is supported by Business Finland
 under Grant No. 1848/31/2015. J. Matas was supported by the OP VVV funded project CZ.02.1.01/0.0/0.0/16\_019/000076 
 Research Center for Informatics.

{\small
\bibliographystyle{ieee}
\bibliography{cvpr2019}
}

\clearpage

\end{document}